\documentclass[conference]{IEEEtran}
\IEEEoverridecommandlockouts
\usepackage{cite}
\usepackage{amsmath,amssymb,amsfonts}
\usepackage{algorithmic}
\usepackage{graphicx}
\usepackage{textcomp}
\usepackage{xcolor}
\def\BibTeX{{\rm B\kern-.05em{\sc i\kern-.025em b}\kern-.08em
    T\kern-.1667em\lower.7ex\hbox{E}\kern-.125emX}}
    
\usepackage{caption}
\usepackage{subcaption}
\usepackage{hyperref}
\usepackage{booktabs}
\newcommand{\ra}[1]{\renewcommand{\arraystretch}{#1}}
\usepackage{listings}
\begin{document}

\title{Post-surgical Endometriosis Segmentation in Laparoscopic Videos}

\author{
    \IEEEauthorblockN{Andreas Leibetseder, Klaus Schoeffmann}
    \IEEEauthorblockA{\textit{Institute of Information Technology} \\
    \textit{Klagenfurt University}\\
    Klagenfurt, Austria \\
    {[aleibets,ks]@itec.aau.at}}
\and
    \IEEEauthorblockN{Jörg Keckstein}
    \IEEEauthorblockA{\textit{Medical Faculty} \\
    \textit{Ulm University}\\
    Ulm, Germany \\
    joerg@keckstein.at}

\and
    \IEEEauthorblockN{Simon Keckstein}
    \IEEEauthorblockA{\textit{University Hospital} \\
    \textit{Ludwig-Maximilians-University}\\
    Munich, Germany \\
    simon.keckstein@med.uni-muenchen.de}
}

\maketitle

\begin{abstract}
Endometriosis is a common women's condition exhibiting a manifold visual appearance in various body-internal locations. Having such properties makes its identification very difficult and error-prone, at least for laymen and non-specialized medical practitioners. In an attempt to provide assistance to gynecologic physicians treating endometriosis, this demo paper describes a system that is trained to segment one frequently occurring visual appearance of endometriosis, namely dark endometrial implants. The system is capable of analyzing laparoscopic surgery videos, annotating identified implant regions with multi-colored overlays and displaying a detection summary for improved video browsing. 
\end{abstract}




\section{Introduction}

Endoscopic surgical procedures are well established particularly in gynecology. The exact diagnosis of various diseases takes place via an endoscopy camera system which is inserted into the abdominal cavity through a small port. The endoscopic image is made available to the surgeon on monitors. The exploration of the abdominal cavity and especially the inner genital tract is very informative and helpful for a correct diagnosis and therapy in the case of painful conditions or pathological findings. One condition commonly treated this way is termed \textit{endometriosis}, which refers to the abnormal growth of uterine-like tissue outside of the uterus and is diagnosed among women of child-bearing age. Affected patients exhibit lesions of varying severity -- often in various locations. Complete identification and recording of all foci and their therapy (removal) is essential for improving symptoms and quality of life of the patient. There are two mainly used systems to classify the disease, the revised American Society for Reproductive Medicine (\textit{rASRM}) score~\cite{canis1997revised} and the \textit{Enzian} classification~\cite{keckstein2003enzian, keckstein2020classification}. The rASRM classification is particularly applicable to the recording of all intraperitoneal lesions, whereas the Enzian classification covers deep endometriosis. The classification is primarily carried out by the surgeon's visual assessment complimenting each other for quantifying a patient's overall condition.

The entire detection of the endometriosis in the partially inaccessible area of the pelvis and the large area of the peritoneum can be limited, and is made more difficult by the different color and appearance of the respective endometrial lesions. Due to these various manifestations of endometriosis, good training and great attention is required from the surgeon during diagnosis. The lack of experience, possibly combined with time pressure under a large operation list, carries the risk of incomplete recording of the disease. This has an essential consequence for the further treatment and the patient's well-being. There is a requirement to prevent misdiagnosis of the disease as far as possible and at the same time to intensify the visual perception of all lesions, especially for doctors in training. This could be supported intra- or post-operatively with the help of image segmentation. 

\begin{figure*}[tbp!]
    \centering
    \begin{subfigure}{.45\textwidth}
      \centering
      \includegraphics[width=.9\linewidth]{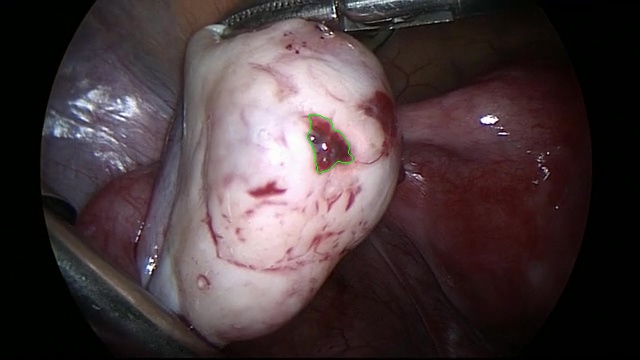}
      \caption{}
      \label{fig:implants_sub1}
    \end{subfigure}%
    \begin{subfigure}{.45\textwidth}
      \centering
      \includegraphics[width=.9\linewidth]{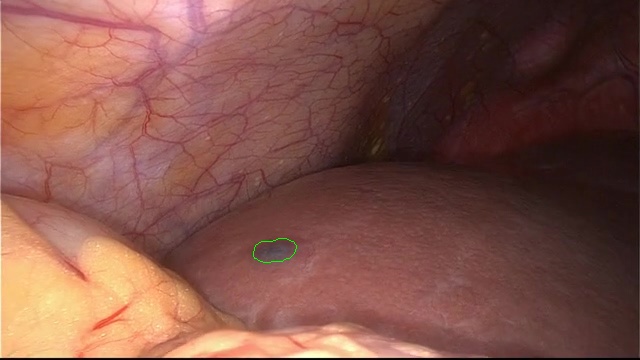}
      \caption{}
      \label{fig:implants_sub2}
    \end{subfigure}\\[1ex]
    \centering
    \begin{subfigure}{.45\textwidth}
      \centering
      \includegraphics[width=.9\linewidth]{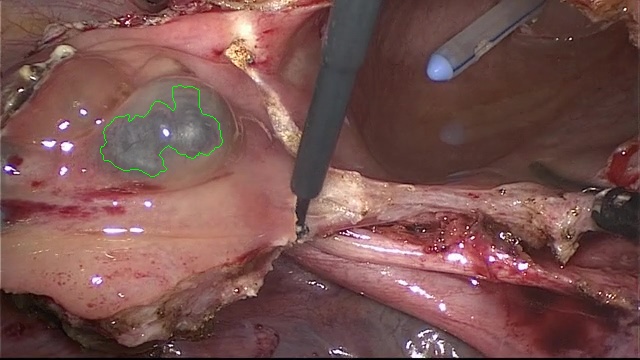}
      \caption{}
      \label{fig:implants_sub3}
    \end{subfigure}
    \begin{subfigure}{.45\textwidth}
      \centering
      \includegraphics[width=.9\linewidth]{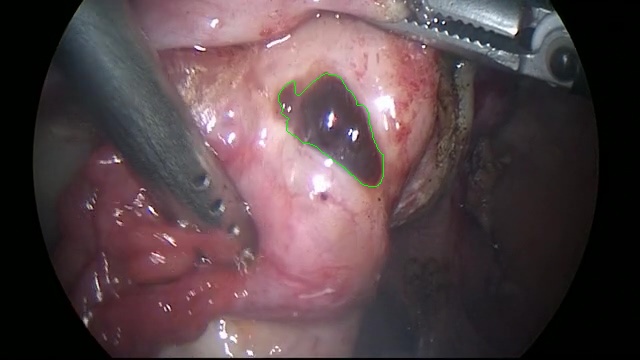}
      \caption{}
      \label{fig:implants_sub4}
    \end{subfigure}
    \caption{Examples of dark endometrial implants}
    \label{fig:implants}
\end{figure*}

With deep learning already heavily employed in medical imaging, it naturally could be regarded as an opportunity for not only improving aforementioned educational training but as well facilitate post-surgical analysis. In order to demonstrate the feasibility of such a goal, for this work we focus on the object segmentation of a specific visual appearance of endometriosis -- dark \textit{endometrial implants}. Figure~\ref{fig:implants} depicts four examples taken from a custom-created ground truth dataset\footnote{\url{https://tinyurl.com/ENIDDS}} including region-based annotations of such pathological areas. When regarding these annotations, it can be observed that, although the indicated regions appear distinctly different from their immediate surroundings, they seem quite similar to other non-pathological areas such as spots of blood or dark vessels. The dataset exclusively contains single-class implant annotations and is used to adapt and train the state-of-the-art deep object segmentation network Mask R-CNN~\cite{DBLP:journals/pami/HeGDG20}, which is a region-based convolutional neural network capable of producing pixel masks for detected objects in addition to bounding boxes generated by an incorporated region proposal network (c.f. Faster R-CNN\cite{ren2015faster}). Overall, we formulate our contributions as follows:

\begin{itemize}
    \item Adapting Mask R-CNN and providing a model for binary segmentation of endometrial implants.
    \item Local and temporal visualization of endometrial implants in laparoscopic surgery videos.
    \item Providing the tool source code as well as pre-trained models for academic purposes\footnote{\url{https://tinyurl.com/EndoSegTool}}.
\end{itemize}

This demonstration highlights partial results of an ongoing more thorough study on the subject of endometriosis segmentation. As such, the following sections intentionally focus on describing the tool and its features rather than portraying the dataset creation and training approach in very much detail.

\section{Endometriosis Segmentation Tool}

The endometriosis segmentation tool can generally be described as an ensemble of technologies combined, resulting in a series of scripts for analyzing post-surgical video archives. These scripts are used for creating annotated output videos as well as a configurable amount of metadata, which can for instance be incorporated into potential interactive systems. As mentioned above, this demo should be regarded as a showcase for highlighting the feasibility of endometriosis segmentation, therefore, we reserve building a fully-fledged user interface for future versions of the tool. In the following sections we describe its architecture, usage, hardware-specific runtime analysis and implementation details.

\subsection{Architecture}

The system's overall architecture is comprised of three three main steps: dataset creation, model training and video analysis (model application). 

We custom-create a single-class lesion dataset from refining parts of the more extensive and multi-class Gynecologic Laparoscopy Endometriosis Dataset~\cite{DBLP:conf/mmm/LeibetsederKSKK20} (GLENDA). The collected base dataset comprises over 350 region-based endometrial implant annotations for 160 frames taken from more than 100 patient cases exhibiting endometriosis. In order to improve the trained segmentation model, we augment this dataset by applying various techniques including rotating, blurring, perspective transformation, desaturation as well as object tracking. For the subsequent training step we divide these various resulting datasets into two different subsets used for training, validation and testing.

\begin{figure}[bp!]
    \centerline{\includegraphics[width=.95\linewidth]{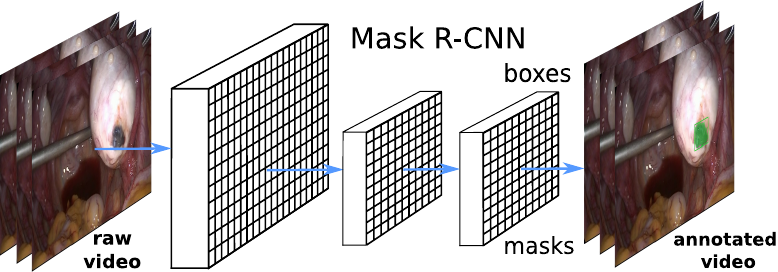}}
    \caption{Video Processing Pipeline.}
    \label{fig:processing}
\end{figure}

\begin{figure*}[tbp!]
    \centering
    \begin{subfigure}{.45\textwidth}
      \centering
      \includegraphics[width=.9\linewidth]{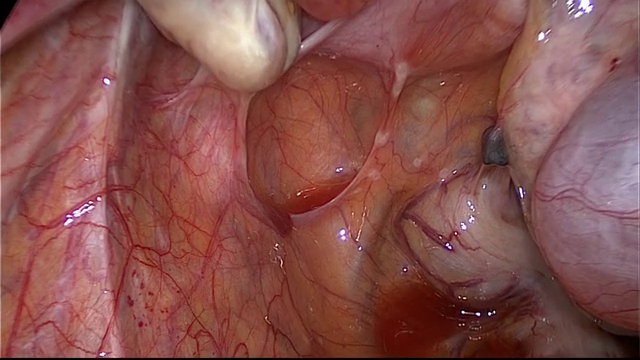}
      \caption{}
      \label{fig:interface_sub1}
    \end{subfigure}%
    \begin{subfigure}{.45\textwidth}
      \centering
      \includegraphics[width=.9\linewidth]{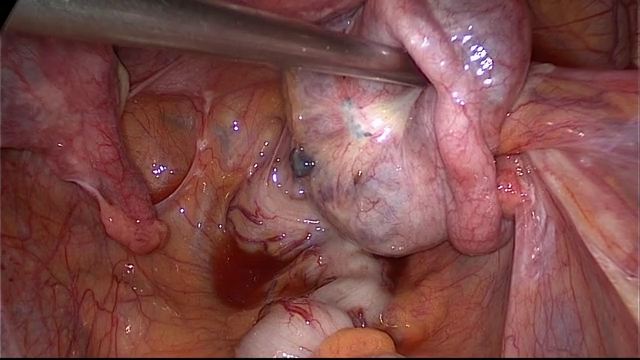}
      \caption{}
      \label{fig:interface_sub2}
    \end{subfigure}
    \centering
    \begin{subfigure}{.45\textwidth}
      \centering
      \includegraphics[width=.9\linewidth]{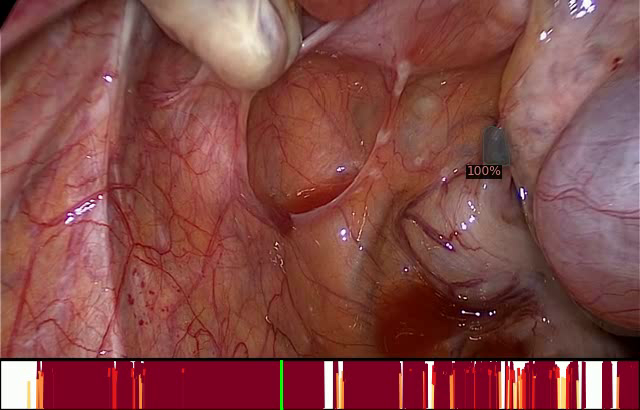}
      \caption{}
      \label{fig:interface_sub3}
    \end{subfigure}
    \begin{subfigure}{.45\textwidth}
      \centering
      \includegraphics[width=.9\linewidth]{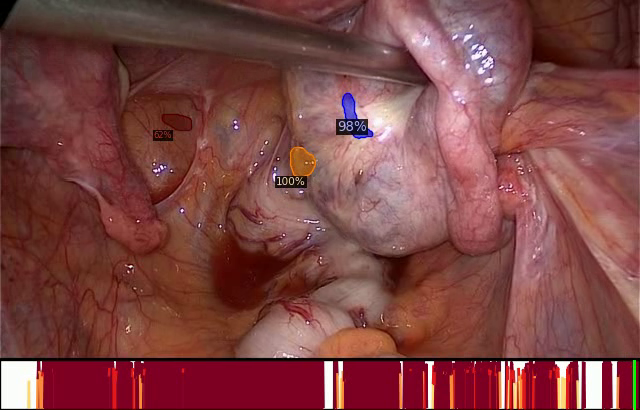}
      \caption{}
      \label{fig:interface_sub4}
    \end{subfigure}
    \caption{Video at two different points in time -- raw (top row) and analyzed (bottom row)}
    \label{fig:interface}
\end{figure*}

As mentioned above, for model training we adapt state-of-the-art object segmentation network MASK R-CNN for transfer learning a single output label. As a backbone network we employ ResNet-101~\cite{he2016deep} together with overall multi-task loss function incorporating class (log loss), bounding box (smooth $L_1$ loss) and mask segmentation (binary cross entropy loss) predictions as described in~\cite{girshick2015fast, DBLP:journals/pami/HeGDG20}. Training is conducted for 50 epochs using a learning rate of $0.001$ and stochastic gradient descent as an optimizer. The best performing model in terms of mean average precision (mAP) for mask segmentation as employed in the MS COCO-detection~\cite{lin2014microsoft} evalutaions is achieved after 29 epochs using rotation as well as cropping for augmentation: 0.642 mAP@0.50IoU at a threshold of 0.5 mask overlap (0.324 mAP for a threshold range of 0.50 to 0.95 with 0.05 steps). This model together with other well-performing models from both splits are made available for download\footnote{\url{https://tinyurl.com/ENIDDS}}.

Finally, we utilize such a model in our system for detecting pathologically suspicious regions with a confidence threshold of 0.50 or above. The employed core processing pipeline is depicted in Figure~\ref{fig:processing}: first a user provides the tool with a raw surgery video, which then is analyzed frame by frame extracting bounding boxes, masks and labels. Whenever results are found, the tool uses the determined segmentation masks to produce annotated frames as well as an overall detection summary in form of an indication bar, as depicted in Figure~\ref{fig:interface}. This bar indicates frame-by-frame detections over-time, colored by detection confidence (yellow to dark red) -- values for multiple detections are averaged. Both, segmentation results as well as indication bar are integrated into the final video output, while additionally marking the current video position with a green horizontal bar. This way, viewers of such annotated output videos at any point in time are provided with an overview of potentially important sections. All extracted data can additionally be stored in JSON-format, as to facilitate the integration in to future interactive video browsing systems. 

\subsection{Hardware and Runtime Analysis}

For implementation, training and evaluation we used a workstation with the following specifications: Intel Core i7-5820K CPU @ 3.30GHz x 6, 32 GiB DDR3 @ 1333 MHz, Nvidia GeForce GTX 1080. On such a machine, model training required approximately 2h to complete. The tool has been implemented using Linux Ubuntu 18.x, but also successfully tested on Windows 10 systems. Given the exclusive utilization of cross-platform technologies (c.f. Section\ref{sec:installation}), it is assumed to be compatible with MacOS as well.

\begin{table}[tbp!]
    \centering
    \ra{1.3}
    \caption{Processing time comparison of 16:9 resolutions.}
    \label{tab:runtime_ms}
    \begin{tabular}{@{}rc@{}}\toprule
        \multicolumn{1}{c}{\textbf{resolution}} & \multicolumn{1}{c}{\textbf{avg in ms}} \\
        \midrule
        $640\times360$  & 153 \\
        $1280\times720$ & 158 \\
        $1920\times1080$ & 170 \\
        $3840\times2160$ & 207 \\
        \bottomrule
    \end{tabular}
\end{table}

Concerning runtime performance, when using GPU processing the system requires an average of approximately 150-250ms of processing time per frame for most videos, as is outlined by Table~\ref{tab:runtime_ms}. Albeit clearly growing with larger resolutions, the processing time essentially depends on resizing the input images, since the generated model's input is resized to fit a restricted distinct pixel range, i.e. 800 pixels for the shortest and 1333 pixels for the longest image side. Hence, assuming a per-frame performance of 170ms we can approximately estimate the overall time requirements of processing an hour of video produced by an endoscope recording in HD resolution with 25 frames per second: $\frac{170\times 25\times60\times60}{1000} = 15 300s = 4h 15m$. 

\subsection{Installation and Usage}
\label{sec:installation}

The tool requires working installations OpenCV\footnote{OpenCV 4.x, \url{https://opencv.org}}, Python 3.x~\footnote{Python 3.x, \url{https://www.python.org}}, FFmpeg\footnote{\url{https://ffmpeg.org}} and Detectron2\footnote{\url{https://github.com/facebookresearch/detectron2}}. All further requirements can simply be installed by running:

\begin{lstlisting}[language=bash]
$ pip install requirements.txt 
\end{lstlisting}

In its most basic use case -- analyzing a single video -- the tool can be executed by running:
    
\begin{lstlisting}[language=bash]
$ python demo.py -i <video file> -m <model file> -o <output folder>
\end{lstlisting}

The tool is also capable of multi-video and -model processing and a detailed description of all available options can be produced by running the script with the '\texttt{-h}' flag. 

\section{Conclusion}

We present a tool for segmenting and annotating endometrial implants in laparoscopic videos. Approaching this problem by combining video object tracking in combination with state-of-the-art image segmentation, we achieve qualitatively good results that can be regarded as a first step towards an interactive post-surgical video archive browser, which could be of great assistance for treatment planning as well as clinical education. Finally, this work represents valuable insights into the feasibility of applying traditional machine learning developed real-world object detection to a practical medical use case.


\section*{Acknowledgments}
This work was funded by the FWF Austrian Science Fund under grant P 32010-N38.

\bibliographystyle{IEEEtran} 
\bibliography{bib} 

\end{document}